%% file: main.tex
\documentclass[10pt,twocolumn]{article}
\usepackage[T1]{fontenc}
\usepackage[margin=0.8in]{geometry}
\usepackage{amsmath,amssymb,graphicx,booktabs}
\usepackage[numbers,sort&compress]{natbib}
\usepackage[hidelinks]{hyperref}
\newcommand{\sg}{\operatorname{sg}}
\newcommand{\E}{\mathbb{E}}
\title{QAMM: Adjoint MeanFlow Matching for\\Few-Step Offline Reinforcement Learning}
\author{\begin{tabular}{ccc}
\textbf{Yuehu Gong}$^{1}$ & \textbf{Shutong Ding}$^{2}$ & \textbf{Mokai Pan}$^{2}$\\
{\footnotesize\texttt{26210980150@m.fudan.edu.cn}} &
{\footnotesize\texttt{dingsht@shanghaitech.edu.cn}} &
{\footnotesize\texttt{panmk2025@shanghaitech.edu.cn}}\\[10pt]
\textbf{Yimiao Zhou}$^{2}$ & \textbf{Jiashu Hou}$^{3}$ &
\textbf{Ye Shi}$^{2}$\\
{\footnotesize\texttt{zhouym2025@shanghaitech.edu.cn}} &
{\footnotesize\texttt{jiashuhou@sjtu.edu.cn}} &
{\footnotesize\texttt{shiye@shanghaitech.edu.cn}}\\[10pt]
\multicolumn{3}{c}{\textbf{Yanwei Fu}$^{1,3}$}\\
\multicolumn{3}{c}{{\footnotesize\texttt{yanweifu@fudan.edu.cn}}}
\end{tabular}\\[8pt]
{\small $^{1}$School of Data Science, Fudan University \quad
$^{2}$ShanghaiTech University}\\
{\small $^{3}$Shanghai Innovation Institute}}
\date{}
\begin{document}
\raggedbottom
\maketitle
\begin{abstract}
\input{sections/abstract}
\end{abstract}
\input{sections/introduction}
\input{sections/related_work}
\input{sections/preliminaries}
\input{sections/method}
\input{sections/experiments}
\input{sections/conclusion}
\section*{Acknowledgments}
We thank Yonghoon Dong for sharing the offline-to-online OGBench evaluation
logs for TRQAM and its baselines used in our comparisons.
\bibliographystyle{plainnat}
\bibliography{references}
\clearpage
\appendix
\input{sections/appendix}

\end{document}

%% file: sections/abstract.tex
Flow policies can model rich action distributions, but their iterative sampling
limits decision speed. Q-learning with Adjoint Matching (QAM) converts the
critic's terminal action gradient into instantaneous flow-velocity targets
without backpropagating through the sampling trajectory. Its policy still
requires numerical integration to generate actions. We propose
\emph{QAMM}, a method that turns the critic-derived adjoint signal into
supervision for MeanFlow's average velocity. The resulting policy learns
finite-interval transport directly and generates actions with few network
evaluations. We derive the adjoint MeanFlow target, specify its gradient
boundaries, and train it with an offline actor--critic. On ten HumanoidMaze
tasks, QAMM produces effective two-call policies and achieves competitive
performance against strong flow-policy baselines. These results show that
adjoint-based Q optimization can be combined with average-velocity learning
to obtain expressive offline policies with few-step action generation.

%% file: sections/introduction.tex
\section{Introduction}
\label{sec:introduction}
\begin{figure*}[t]
\centering
\includegraphics[width=\textwidth]{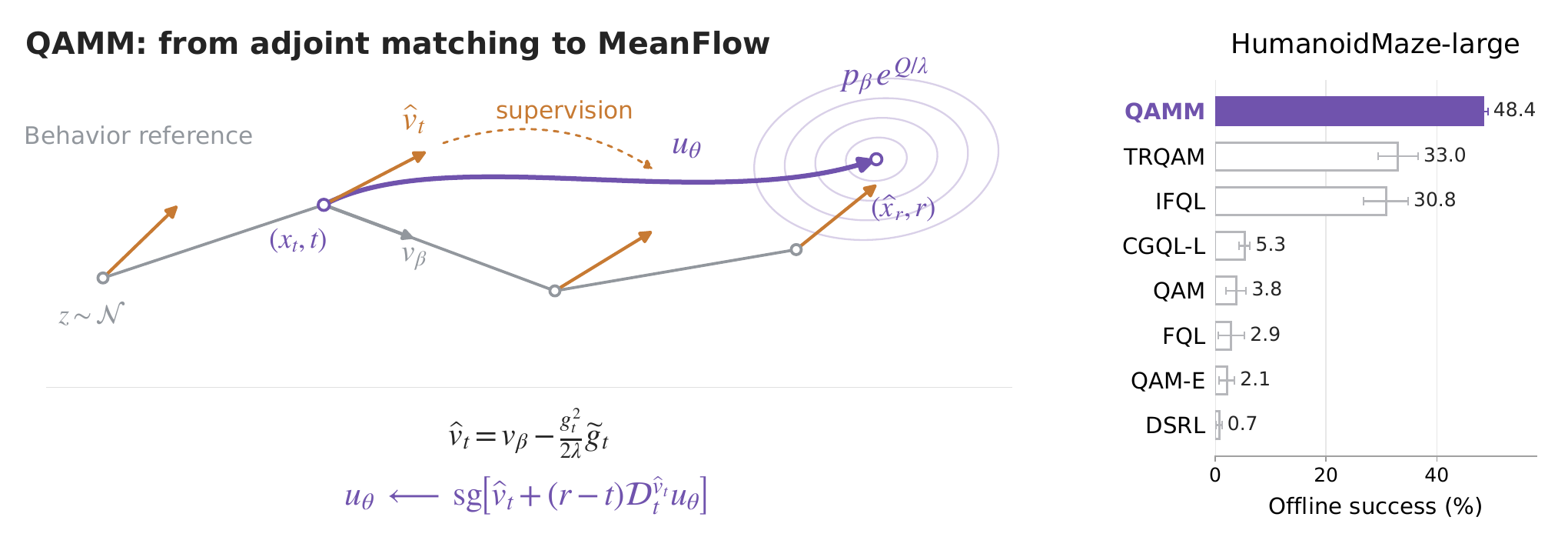}
\caption{QAMM: learning average velocities from critic gradients.
\textbf{Left:} adjoint-corrected velocities (orange) originate at fixed
states on the gray reference path. At the selected $(x_t,t)$, the corrected
velocity supervises MeanFlow through the target below. The purple curve
schematically represents predictions $\widehat x_r=x_t+(r-t)u_\theta(s,x_t,r,t)$
as the destination $r$ varies; the dotted arrow denotes regression supervision.
Contours illustrate the ideal behavior-regularized Q tilt.
\textbf{Right:} offline success on the five HumanoidMaze-large tasks, using
the means and seed standard deviations in Table~\ref{tab:main}. QAMM uses
two seeds and the tuned setting; the supplied baselines use eight seeds.}
\label{fig:teaser}
\end{figure*}

Flow and diffusion policies have become useful action models in offline
reinforcement learning~\citep{wang2022diffusionql,hansenestruch2023idql,zhang2025ewfm,park2025fql,alles2025flowq,li2026qam,dong2026trqam}.
They can fit complex behavior distributions, but policy improvement and action
generation remain coupled to an iterative sampler. Methods that optimize
sampled actions often differentiate through this sampler or use an auxiliary
policy~\citep{park2025fql,espinosadice2025shortcut,chen2025mirror};
value-weighted regression instead discards the critic's action
gradient~\citep{ding2024qvpo,zhang2025ewfm}. Intermediate score
guidance~\citep{psenka2024qsm,fang2025dac} uses the critic gradient at noisy
actions, which need not agree with its gradient at the final action.
Q-learning with Adjoint Matching (QAM)~\citep{li2026qam} addresses this
optimization problem: a lean adjoint converts the terminal Q gradient into velocity targets
along a trajectory, avoiding backpropagation through the sampling chain.
Its learned velocity is instantaneous, so accurate action generation still
requires integrating a flow.

Flow matching~\citep{lipman2023flow,liu2023rectified} traditionally learns an
instantaneous velocity. MeanFlow~\citep{geng2025meanflows} provides a different parameterization:
its network predicts the average velocity between two times and can therefore
represent a long transport step directly. Applying QAM to a MeanFlow network
requires more than changing the network input. The adjoint label must supervise
both the instantaneous term of the MeanFlow identity and the direction of its
derivative. This raises a concrete training question: can we use Q's action
gradient to learn a finite-interval flow without differentiating through the
policy's full sampling trajectory?

We propose \emph{QAMM}, Adjoint MeanFlow Matching. We propagate the critic's
terminal gradient backwards to construct an adjoint-corrected velocity, then
use that velocity in a stop-gradient average-velocity target. The learned field
supports short generation intervals and a two-call action sampler. For offline
RL, we initialize from a behavior flow, jointly train the actor and critic,
and construct training trajectories by noising policy-generated endpoints.
We use an adapted path-space controller~\citep{dong2026trqam} to stabilize Q
guidance. At deployment, two-call refinement with independent re-noising
produces actions, informed by recent work on stochastic generative
control~\citep{pan2025ado}.

Our contributions are threefold. First, we derive an adjoint MeanFlow target
that turns the critic's terminal action gradient into finite-interval
average-velocity supervision, with explicit gradient boundaries. Second, we
build an offline training procedure that couples policy-generated endpoints
to Markov noising trajectories and supervises both diagonal and off-diagonal
time pairs, while retaining two-call action generation at deployment. Third,
we evaluate this policy on five HumanoidMaze-medium and five HumanoidMaze-large
tasks against flow-policy baselines and report initial backup and sampling
diagnostics. Additional domains and seeds will test how broadly the result
extends.

%% file: sections/related_work.tex
\section{Related Work}
\label{sec:related}
\paragraph{Generative policies for offline RL.}
Behavior-constrained offline RL includes Gaussian actors and value-weighted
extraction~\citep{kostrikov2022iql,tarasov2023rebrac}. Diffusion
Q-learning~\citep{wang2022diffusionql} and IDQL~\citep{hansenestruch2023idql}
extend policy expressivity. For diffusion and flow actors, Q can enter through
action or noise-space optimization~\citep{park2025fql,wagenmaker2025latent},
value- or energy-weighted regression~\citep{ding2024qvpo,zhang2025ewfm,alles2025flowq},
or intermediate score guidance~\citep{psenka2024qsm,fang2025dac}.
These uses of Q differ from a terminal action gradient propagated into
average-velocity supervision.

\paragraph{Adjoint guidance and average velocity.}
Adjoint Matching~\citep{domingoenrich2024adjoint} gives the memoryless control
formulation and lean adjoint; QAM~\citep{li2026qam} applies it to learned
critics. Flow matching and rectified flow~\citep{lipman2023flow,liu2023rectified}
provide the instantaneous-velocity background. MeanFlow~\citep{geng2025meanflows} supplies the average-velocity
identity. QAMM combines these ingredients to supervise average velocity with
the critic's adjoint signal. TRQAM~\citep{dong2026trqam} introduces adaptive
path-space control for adjoint learning; we adapt that mechanism for stability.
\paragraph{Few-step generative control.}
Much Ado About Noising~\citep{pan2025ado} studies stochastic iterative
computation and simple two-step policies in robotic control. Our use of
independent re-noising is a sampling choice, whereas the present training
contribution is the Q-adjoint average-velocity target. Consistency
policies~\citep{ding2024consistency}, shortcut models~\citep{espinosadice2025shortcut},
and one-step flow policy mirror descent~\citep{chen2025mirror} also address
sampling cost, with different training objectives. In image and video
generation, MeanFlowNFT~\citep{huang2026meanflownft} also studies reward
optimization for average-velocity models using a different, forward-process
objective. Broader comparisons of scalar-Q and Q-gradient learning are left
for future work.

%% file: sections/preliminaries.tex
\section{Preliminaries}
\label{sec:preliminaries}
\subsection{Offline reinforcement learning}
Consider an MDP with state $s$, continuous action $a$, transition kernel $P$,
reward $R$, and discount $\gamma\in[0,1)$. An offline dataset $\mathcal D$
contains transitions $(s,a,R,s',m)$, where $m$ indicates whether bootstrapping
continues. We seek a policy maximizing expected discounted return without
additional environment interaction during training. A critic ensemble
$\{Q_{\phi_i}\}_{i=1}^M$ is trained by temporal-difference regression; its target
ensemble mean is denoted $\overline Q$. The learned behavior model approximates
dataset actions conditioned on state. Generation times $t,r$ describe internal
action generation and are distinct from environment steps.
\subsection{Average velocity}
Generation proceeds from Gaussian noise at $t=0$ to an action at $t=1$.
We write $u_\theta(s,x,r,t)$ with current time $t$ and destination time $r$,
where $0\leq t\leq r\leq1$. For an exact ODE flow map $\Phi^v$,
\begin{equation}
u^v(s,x,r,t)=\frac{\Phi^v(r;t,x)-x}{r-t},\quad r>t.
\end{equation}
The diagonal limit is $u^v(s,x,t,t)=v(s,x,t)$. Holding $r$ fixed gives
\begin{equation}
u^v=v+(r-t)\mathcal D_t^v u^v,\quad
\mathcal D_t^a u=\partial_tu+(\partial_xu)a.
\end{equation}
\subsection{Memoryless control and lean adjoints}
For independent Gaussian noise $z$ and a behavior action $a$, the interpolation
$x_t=(1-t)z+ta$ has conditional velocity $a-z$. Its marginal velocity defines
an ODE transport. The associated ideal memoryless SDE used in AM/QAM has drift
$2v_\beta-x/t$ and diffusion squared $2(1-t)/t$.
Memorylessness means independence of its initial and terminal variables; it is
not a statement that the intermediate trajectory points are independent.

AM relates a terminal reward to stochastic control with a quadratic drift cost.
The corresponding path-KL interpretation requires matching diffusion and initial
laws and suitable change-of-measure assumptions. Lean adjoints transport a
terminal reward derivative backwards through input Jacobians of the reference
drift. Section~\ref{sec:method} specifies the equations, signs, and scaling used
for QAMM. Singular endpoints are interpreted through the ideal entrance law;
the implementation employs a separate finite-step regularization.

%% file: sections/method.tex
\section{Adjoint MeanFlow Matching}
\label{sec:method}
We learn a state-conditioned average-velocity field $u_\theta(s,x,r,t)$, where
$t$ is the current generation time and $r$ is the destination time. Generation
runs from noise at zero to actions at one. Our central construction converts
critic-derived adjoints into supervision for this field. The behavior model,
trust-region controller, and mixed critic backup support this construction;
their underlying principles are inherited from prior work.

\subsection{From critic gradients to velocity targets}
For a fixed environment state $s$, let $v_\beta$ denote the reference velocity.
The ideal memoryless formulation of AM/QAM~\citep{domingoenrich2024adjoint,li2026qam}
uses
\begin{align}
g_t^2&=2(1-t)/t,\nonumber\\
f_\beta(s,x,t)&=2v_\beta(s,x,t)-x/t,\label{eq:reference}\\
f_\theta(s,x,t)&=2u_\theta(s,x,t,t)-x/t.\nonumber
\end{align}
The corresponding controlled SDE is $dX_t=f_\theta\,dt+g_t\,dB_t$.
For a fixed critic $Q$ and coefficient $\lambda>0$, the associated ideal
KL-regularized control problem is
\begin{equation}
\min_f\;\lambda D_{\mathrm{KL}}(P^f\Vert P^\beta)
-\E_{P^f}[Q(s,X_1)],\label{eq:soc}
\end{equation}
where the path measures have the same initial distribution and diffusion.
The memoryless construction, together with the requisite regularity and
integrability assumptions, yields a terminal tilt proportional to
$p_\beta(a\mid s)\exp(Q(s,a)/\lambda)$. This is a statement about the ideal
control problem, not the exact distribution of our finite-step sampler.

Along a training trajectory, the lean adjoint satisfies
\begin{align}
\dot{\widetilde g}_t
&=-[\partial_x f_\beta(s,X_t,t)]^\top\widetilde g_t,
&\widetilde g_1&=-\nabla_x Q(s,X_1).\label{eq:adjoint}
\end{align}
Following the trust-region scaling of TRQAM~\citep{dong2026trqam}, it gives
the instantaneous-velocity label
\begin{equation}
\widehat v_t=\sg\!\left[v_\beta(s,X_t,t)
-\frac{g_t^2}{2\lambda}\widetilde g_t\right].\label{eq:vhat}
\end{equation}

\paragraph{Implemented adjoint.}
We use the mean target critic $\overline Q$ and differentiate
$Q_c(s,x)=\overline Q(s,\operatorname{clip}(x,-1,1))$ at the generated endpoint.
Thus the terminal derivative includes the action-clipping Jacobian.
With $h=1/K$ and $t_k=kh$, the implementation propagates the unscaled adjoint:
\begin{equation}
\widetilde g_k=\widetilde g_{k+1}
+h[\partial_x f_\beta(s,X_{t_k},t_{k+1})]^\top
\widetilde g_{k+1}.\label{eq:discrete-adjoint}
\end{equation}
It replaces $g_{t_k}^2$ in Eq.~\eqref{eq:vhat} with
$g_{h,k}^2=2(1-t_k+h)/(t_k+h)$. These choices regularize the singular
endpoint and discretize the adjoint; they are not exact continuous-time identities.
Reference parameters are fixed within this computation, while their input
Jacobians remain available for vector--Jacobian products.

\subsection{Average-velocity supervision}
MeanFlow's identity~\citep{geng2025meanflows} relates an average velocity to an
instantaneous velocity through a material derivative with the destination $r$
held fixed. We substitute the adjoint label in both appearances of that velocity:
\begin{align}
T_\theta(s,X_t,r,t)
&=\sg\!\left[\widehat v_t+(r-t)\mathcal D_t^{\widehat v_t}u_\theta\right],
\label{eq:qamm-target}\\
\mathcal D_t^{\widehat v_t}u_\theta
&=\partial_tu_\theta+(\partial_xu_\theta)\widehat v_t.\nonumber
\end{align}
This derivative is evaluated by a JVP on $(x,t)$ with direction
$(\widehat v_t,1)$. The live student supplies the differentiated field; the
reference behavior network is not used as a fixed derivative teacher.
We stop parameter gradients through the sampled path, the adjoint, the velocity
label, and the entire target. Consequently, optimization differentiates only
the prediction side, without backpropagating through the sampling or adjoint chain.

\paragraph{Three destination times.}
For every trajectory point, we evaluate three equally weighted terms:
\begin{equation}
r_0=t,\qquad r_1=1,\qquad
r_2=t+(1-t)U,\quad U\sim\mathcal U(0,1).\label{eq:three-r}
\end{equation}
The diagonal term reduces exactly to $\widehat v_t$ because its derivative
coefficient vanishes. The other terms supervise terminal and intermediate
displacements. Even at a sample with $r=t$, $r$ is held fixed during the JVP.
Our raw time input is $(t,r-t)$; the JVP therefore also differentiates the
interval feature with respect to $t$. Actor times come from the $K$-point
trajectory grid, rather than an additional independent draw of $t$.

The basic regression objective averages
$\|u_\theta-T_\theta\|_2^2$ over sampled states and time pairs. Our
implementation applies a residual-dependent weight for numerical stability;
the exact objective and its theoretical consequence are given with the other
practical choices below.

\paragraph{Conditional interpretation.}
For a fixed differentiable student, suppose that under the specified path
measure $\E[\widehat v_t\mid X_t=x]=v^*(s,x,t)$ and that the destination draw
is conditionally independent of the label noise. Linearity in the JVP direction
then gives the numerical target identity
\begin{equation}
\E[T_\theta\mid X_t=x,r,t]
=v^*+(r-t)\mathcal D_t^{v^*}u_\theta.\label{eq:conditional}
\end{equation}
Thus, under these assumptions, the unweighted population target is compatible
with the MeanFlow field induced by $v^*$. This argument does not establish the
conditional-mean premise for arbitrary training paths, or convergence of student
updates. Approximate references, discretization, moving critics, and the
practical residual weighting fall outside this identity's regression conclusion.
The assumptions and the scope of this conditional calculation are detailed in
Appendix~\ref{app:proofs}.

We pretrain a behavior MeanFlow on dataset actions and initialize the policy
from it. Following the joint actor--critic training protocol of
QAM/TRQAM~\citep{li2026qam,dong2026trqam}, we continue updating the behavior
model during offline learning. Its EMA supplies the adjoint reference
$v_\beta(s,x,t)=u_{\bar\beta}(s,x,t,t)$, while behavior actions used in critic
backups use the two-call sampler. The behavior regression target and update
details are given in Appendix~\ref{app:behavior}.

\subsection{Training paths and few-step deployment}
Define the endpoint map
\begin{equation}
C_\theta(s,x,t)=x+(1-t)u_\theta(s,x,1,t).
\end{equation}
Our two-call sampler draws independent standard Gaussian $z_0,z_1$ and computes
\begin{align}
y^{(0)}&=C_\theta(s,z_0,0),\nonumber\\
x_{1/2}^{\mathrm{ref}}&=\tfrac12 y^{(0)}+\tfrac12 z_1,\nonumber\\
y^{(1)}&=C_\theta(s,x_{1/2}^{\mathrm{ref}},\tfrac12).\label{eq:refinement}
\end{align}
Deployment returns $\operatorname{clip}(y^{(1)},-1,1)$. This is endpoint
refinement using fresh noise, not two sequential MeanFlow intervals or two Euler
steps. The current policy also uses this sampler for critic backups.

\paragraph{Endpoint-noised training path.}
The main configuration first generates the un-clipped endpoint $X_1=y^{(1)}$.
It then constructs a complete trajectory backwards on the $K=10$ time grid:
\begin{equation}
X_\tau=\frac{\tau}{t}X_t+
\sqrt{(1-\tau)^2-\frac{\tau^2}{t^2}(1-t)^2}\,\xi_{\tau,t},\label{eq:bridge}
\end{equation}
for consecutive $0\leq\tau<t\leq1$, with independent Gaussian noise at every
transition. In the ideal memoryless reference, this is the reverse conditional
Markov kernel. It cannot be replaced by independently sampled time marginals
or by a straight line with shared noise, since the adjoint depends on the joint
path. The initial state reached by this backward chain need not equal $z_0$.
For a learned approximate reference, equivalence to its actual conditional path
law requires additional justification.

We detach this training path before actor regression. Constructing it still
requires multiple transitions even though deployment uses only two network
calls.

\subsection{Practical stabilization and critic learning}
\paragraph{Residual weighting.}
Following the residual-adaptive weighting used in MeanFlow
training~\citep{geng2025meanflows,geng2025improvedmeanflows}, for action
dimension $d$, write
$e_{k,j}=\|u_\theta-T_\theta\|_2^2/d$. The actual actor objective is
\begin{align}
w(e)&=\operatorname{clip}((e+c)^{-p},10^{-6},10^6),\nonumber\\
L_{\mathrm{QAMM}}&=\operatorname{mean}_{b,k,j}
\left[\sg(w(e_{k,j}))e_{k,j}\right].\label{eq:weighted-loss}
\end{align}
The experiments use $p=0.3$, $c=10^{-3}$, no additional time weighting,
and no target clipping. We also record the unweighted mean regression error.
Since $w$ depends on the residual, the practical optimum need not equal the
conditional mean of the unweighted target.

\paragraph{Adapted path-space controller.}
For matched ideal SDE path measures satisfying Girsanov's conditions,
TRQAM's path-space relation can be expressed in velocity coordinates as
\begin{equation}
D_{\mathrm{KL}}(P^\theta\Vert P^\beta)
=\E_{P^\theta}\int_0^1\frac{2\|v_\theta-v_\beta\|_2^2}{g_t^2}\,dt.
\end{equation}
Our implemented diagnostic is
\begin{equation}
\widehat D=\operatorname{mean}_b\sum_k\frac{2h}{g_{h,k}^2}
\|u_\theta(s,X_k,t_k,t_k)-v_\beta(s,X_k,t_k)\|_2^2.
\end{equation}
The experiments considered here use action horizon one; for action chunks, the
implementation additionally divides this quantity by the horizon.
Because the main endpoint-noised path is not established to be the path law of
the diagonal student drift, we call $\widehat D$ a \emph{path-KL surrogate}.
It is not a certified KL bound for the deployed two-call policy.

Following the TRQAM-inspired feedback mechanism, let $l$ be the internal
controller variable and $\lambda=\kappa l$ the effective guidance coefficient:
\begin{align}
\overline D_n&=(1-\rho)\overline D_{n-1}
+\rho\min(\widehat D_n,c_{\mathrm{KL}}\epsilon),\nonumber\\
l_{n+1}&=\operatorname{clip}\left(l_n+\eta(\overline D_n-\epsilon),
l_{\min},l_{\max}\right).\label{eq:controller}
\end{align}
Clipping is applied to the batch-mean statistic before the EMA. Larger
$\lambda$ reduces the adjoint correction without changing the noising kernel.
The center uses $\kappa=3$, effective $\lambda_0=3$, $\epsilon=0.5$,
$\rho=0.1$, $\eta=0.01$, and $c_{\mathrm{KL}}=2$.
Surrogate tracking does not establish convergence or exact feasibility.

\paragraph{Fixed behavior/current-policy backup.}
For a transition $(s,a,R,s',m)$ with continuation mask $m$, sample
$a_\theta$ from the current two-call policy and $a_\beta$ from the same refinement
procedure applied to the EMA behavior MeanFlow. With a fixed mixture coefficient
$\alpha$, the horizon-one critic target is
\begin{align}
V_{\mathrm{mix}}(s')&=(1-\alpha)\overline Q(s',a_\theta)
+\alpha\overline Q(s',a_\beta),\nonumber\\
y_Q&=\sg\!\left[R+\gamma m V_{\mathrm{mix}}(s')\right].\label{eq:backup}
\end{align}
Each branch uses one sampled action, clipped to the action bounds. We regress
the critic ensemble against $y_Q$ using the dataset validity mask. The center
uses ten critics, ensemble means without an uncertainty penalty, and
$\alpha=0.25$. Mixing occurs between Q values, not between action vectors.
For $\alpha>0$ this is a mixed-policy backup, not pure current-policy evaluation.
The component is practical and not claimed as a novel Bellman operator;
dynamic mixture coefficients are outside the present method.

%% file: sections/experiments.tex
\begin{table*}[t]
\centering\scriptsize
\caption{Offline success rate (\%, mean $\pm$ seed standard deviation),
averaged over 800k--1M updates. QAMM uses two seeds, other methods eight.
Large QAMM uses the tuned setting, selected on one reported seed. Domain
means weight five tasks equally. Bold marks the highest and underline the
second-highest unrounded mean in each row.}
\label{tab:main}
\resizebox{\textwidth}{!}{\begin{tabular}{llrrrrrrrr}
\toprule
Dataset & Task & FQL & CGQL-L & DSRL & IFQL & QAM & QAM-E & TRQAM & QAMM\\
\midrule
Medium & 1 & $82.5\pm13.4$ & $81.2\pm4.3$ & $15.6\pm14.0$ & $\boldsymbol{91.0\pm1.4}$ & $31.9\pm24.8$ & $20.7\pm19.5$ & $\underline{86.8\pm2.4}$ & $67.8\pm3.0$\\
 & 2 & $98.8\pm0.8$ & $\boldsymbol{99.2\pm0.4}$ & $84.6\pm5.5$ & $94.0\pm1.6$ & $\underline{99.2\pm0.8}$ & $98.2\pm0.8$ & $87.7\pm2.5$ & $81.6\pm6.8$\\
 & 3 & $80.2\pm14.5$ & $0.1\pm0.1$ & $57.6\pm22.6$ & $\boldsymbol{94.6\pm1.3}$ & $89.3\pm5.1$ & $70.9\pm21.3$ & $88.4\pm1.6$ & $\underline{90.0\pm1.2}$\\
 & 4 & $0.0\pm0.0$ & $0.0\pm0.0$ & $0.0\pm0.0$ & $\boldsymbol{81.4\pm2.5}$ & $0.0\pm0.0$ & $0.0\pm0.0$ & $58.9\pm9.7$ & $\underline{60.2\pm7.4}$\\
 & 5 & $97.6\pm3.2$ & $\underline{98.8\pm0.7}$ & $85.5\pm2.3$ & $98.2\pm1.0$ & $\boldsymbol{98.8\pm0.6}$ & $98.5\pm1.0$ & $96.2\pm2.3$ & $93.6\pm2.0$\\
 & Mean & $71.8\pm4.8$ & $55.9\pm0.9$ & $48.6\pm6.2$ & $\boldsymbol{91.8\pm0.6}$ & $63.9\pm5.4$ & $57.7\pm5.2$ & $\underline{83.6\pm2.5}$ & $78.6\pm2.8$\\
\midrule
Large & 1 & $3.3\pm4.5$ & $12.3\pm6.2$ & $0.7\pm0.9$ & $28.6\pm4.9$ & $8.2\pm7.5$ & $4.6\pm4.1$ & $\underline{45.3\pm5.9}$ & $\boldsymbol{69.6\pm3.6}$\\
 & 2 & $0.0\pm0.0$ & $0.0\pm0.0$ & $0.0\pm0.0$ & $\underline{7.6\pm5.0}$ & $0.0\pm0.0$ & $0.0\pm0.0$ & $6.6\pm2.9$ & $\boldsymbol{8.2\pm8.2}$\\
 & 3 & $6.3\pm1.4$ & $13.9\pm5.2$ & $1.1\pm0.9$ & $\boldsymbol{75.0\pm3.3}$ & $5.8\pm1.3$ & $2.7\pm1.4$ & $30.6\pm9.4$ & $\underline{51.8\pm12.6}$\\
 & 4 & $0.1\pm0.2$ & $0.1\pm0.1$ & $0.0\pm0.0$ & $\underline{42.9\pm16.3}$ & $0.0\pm0.0$ & $0.0\pm0.0$ & $38.1\pm9.8$ & $\boldsymbol{59.2\pm4.8}$\\
 & 5 & $4.9\pm11.9$ & $0.1\pm0.1$ & $1.5\pm2.1$ & $0.0\pm0.0$ & $4.8\pm7.1$ & $2.9\pm5.1$ & $\underline{44.5\pm3.0}$ & $\boldsymbol{53.0\pm1.4}$\\
 & Mean & $2.9\pm2.4$ & $5.3\pm1.0$ & $0.7\pm0.5$ & $30.8\pm4.0$ & $3.8\pm1.8$ & $2.1\pm1.4$ & $\underline{33.0\pm3.6}$ & $\boldsymbol{48.4\pm0.8}$\\
\bottomrule
\end{tabular}}
\end{table*}

\begin{figure*}[t]
\centering\includegraphics[width=\textwidth]{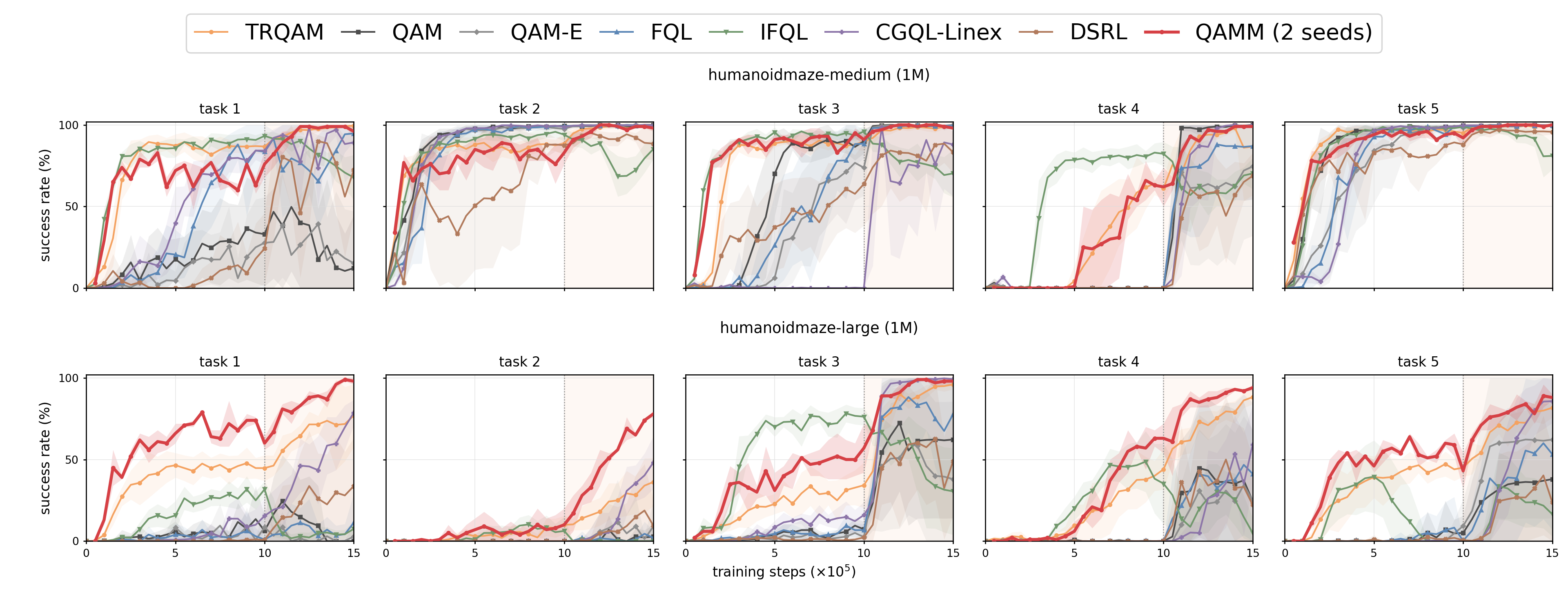}
\caption{Per-task offline-to-online success curves. Columns are tasks 1--5;
rows are HumanoidMaze-medium and HumanoidMaze-large. The vertical line at 1M
updates separates offline and online training. Lines show seed means and
bands show $\pm1$ seed standard deviation. Large QAMM uses the tuned setting.}
\label{fig:curves}
\end{figure*}

\begin{figure*}[t]
\centering\includegraphics[width=\textwidth]{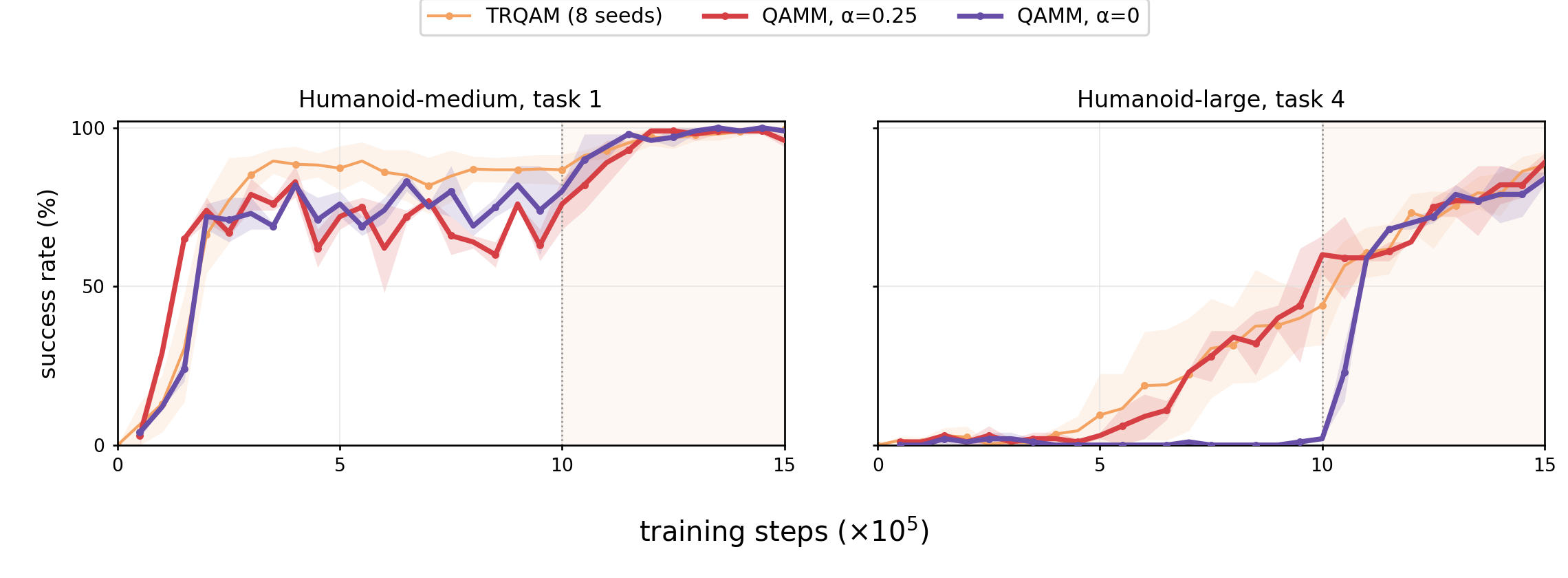}
\caption{Fixed-backup-mixture ablation on HumanoidMaze-medium task 1 and
HumanoidMaze-large task 4. Red uses $\alpha=0.25$ and purple uses $\alpha=0$;
all other center settings and training budgets are matched. Lines and bands
show two-seed mean and $\pm1$ standard deviation. TRQAM's eight-seed curve
is context, not an ablation arm.}
\label{fig:alpha-ablation}
\end{figure*}

\begin{figure*}[t]
\centering
\includegraphics[width=.57\textwidth]{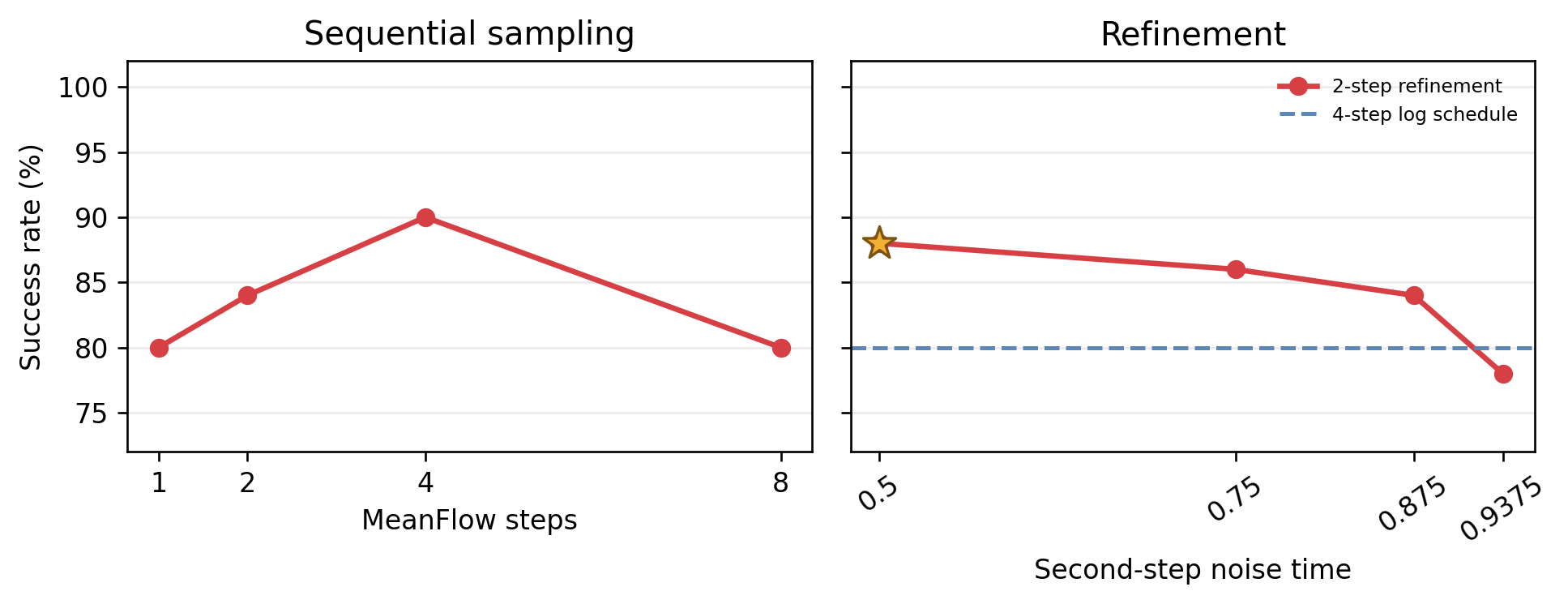}\hfill
\includegraphics[width=.40\textwidth]{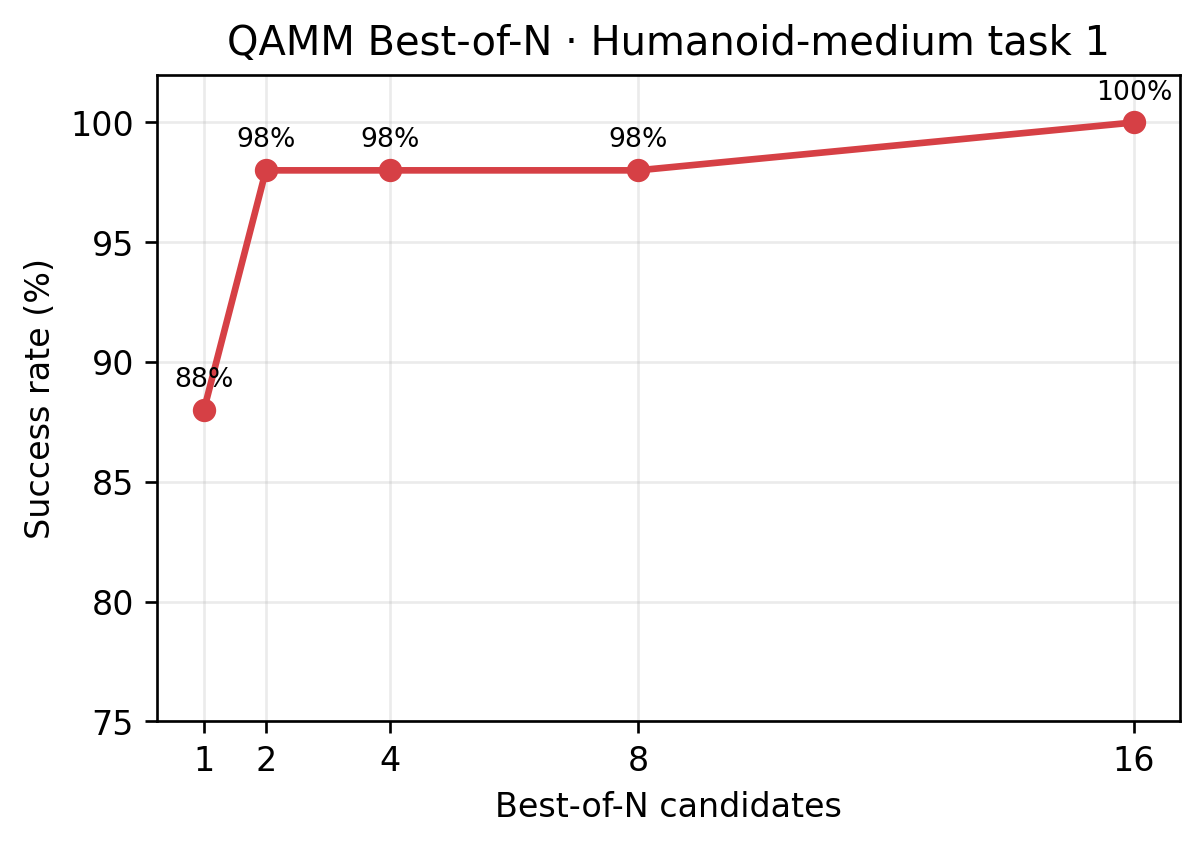}
\caption{Sampling diagnostics on HumanoidMaze-medium task 1 (50 episodes per
setting, one historical stage-1 checkpoint). Left: sequential sampling and
endpoint refinement; the star marks $t=0.5$, QAMM's default two-call sampler.
Right: critic-selected Best-of-$N$ with that sampler. The checkpoint uses a
different backup mixture from the final configuration, so these comparisons
are exploratory.}
\label{fig:sampling-ablation}
\end{figure*}

\section{Experiments}
\label{sec:experiments}
\subsection{Setup}
We study the five tasks each of HumanoidMaze-medium and HumanoidMaze-large in
OGBench~\citep{park2025ogbench}. We compare QAMM's two-call policy with the
published flow-policy baselines using offline success and offline-to-online
learning curves. QAMM has two training seeds and the supplied baseline logs
have eight. The baselines were not re-evaluated under a common harness, so
cross-method differences are descriptive. Training schedules, aggregation,
and hyperparameters are specified in Appendix~\ref{app:config}.

\subsection{HumanoidMaze results}

Table~\ref{tab:main} shows that QAMM learns competitive two-call policies on
Medium and reaches a higher descriptive five-task mean than TRQAM on Large.
Performance is task dependent: the Medium gap is concentrated in tasks 1 and
2, whereas Large gains occur on tasks 1, 3, 4, and 5. Large tasks 4 and 5
share a training goal but have different evaluation initial states; same-seed
training artifacts are therefore dependent. Figure~\ref{fig:curves} shows
further improvement during online continuation. Over the final five online
evaluations, QAMM averages $98.7\%$ on Medium and $87.2\%$ on Large, versus
$97.5\%$ and $71.2\%$ for the supplied TRQAM curves. Unequal seed counts,
selection on a reported seed, and protocol differences preclude a
significance claim.

\subsection{Ablations and sampling analysis}

The Large center averages $39.7\%$ offline and $75.9\%$ online, while the
tuned configuration ($\alpha=0.5$, KL budget $0.75$) averages $48.4\%$ and
$87.2\%$ under the same summary
rule. The per-task results show uneven effects; changing
$\alpha$ and KL together does not isolate either factor. Removing the
behavior component of the mixed Bellman backup changes Medium task 1's
offline mean from $67.8\%$ to $76.0\%$, but Large task 4's from $42.0\%$
to $0.6\%$ under center settings (two seeds each). Figure~\ref{fig:alpha-ablation} shows that the $\alpha=0$ Large task 4
policy stays near zero during offline training and recovers after online
updates, while both Medium task 1 settings reach high online success.
This is task-dependent sensitivity, not a uniform benefit of either
backup choice.

Figure~\ref{fig:sampling-ablation} probes action generation with a single
historical QAMM checkpoint on Medium task 1. Sequential MeanFlow sampling
achieves $80/84/90/80\%$ success with $1/2/4/8$ calls: four calls are best in
this small evaluation, while more calls are not monotonically better.
The four-call logarithmic refinement schedule reaches $80\%$, illustrating
that call count alone does not determine performance. For two-call endpoint
refinement, success changes from $88\%$ at the training
re-noising time $t=0.5$ to $86/84/78\%$ at $t=0.75/0.875/0.9375$.
With critic selection among $N$ independently generated two-call candidates,
Best-of-$N$ yields $88/98/98/98/100\%$ for $N=1/2/4/8/16$. This improvement
uses $2N$ policy calls and additional critic evaluations, so it does not
represent a free gain in few-call deployment. The checkpoint predates the
final backup-mixture choice, and each setting uses only 50 episodes; these
figures motivate a final-checkpoint, multi-seed comparison rather than a
general performance claim.

%% file: sections/conclusion.tex
\section{Discussion and Conclusion}
QAMM connects critic-derived adjoint supervision to an average-velocity policy.
Its target uses the adjoint-corrected velocity both as a regression base and as
the direction of the MeanFlow derivative. Explicit gradient boundaries permit
this training without differentiation through the full sampling trajectory.
The resulting representation supports two-call refinement.

The present evaluation covers two Humanoid datasets and illustrates that
adjoint-trained average velocities can support effective offline control with
few policy calls. Further tasks and independent seeds will clarify the scope
of the empirical result. Few network calls alone do not establish an end-to-end
speedup, which requires a matched latency measurement.

The theory explains a conditional compatibility of the target, not convergence
of the weighted, jointly trained algorithm. The learned reference, finite-step
adjoint, endpoint-noised path law, and deployed refinement introduce distinct
approximations. In particular, the implemented controller tracks a surrogate
rather than certifying the KL of the few-step policy. Clarifying these boundaries
is important when extending adjoint learning to average-velocity models.
The adjoint reference uses only the diagonal behavior velocity
$u_\beta(s,x,t,t)$; the behavior model's finite-interval predictions do not
supervise QAMM at non-diagonal time pairs. Using that information as additional
supervision may improve few-call accuracy, but remains to be tested.
Restricting the destination-time domain remains a possible simplification for
future work, subject to separate accuracy and stability evaluation.

%% file: sections/appendix.tex
\section{Derivations and Conditional Results}
\label{app:proofs}
\subsection{Average-velocity identity in forward generation time}
Fix $s$ and destination $r$. Suppose $v$ generates a unique differentiable flow,
and let $x_t$ follow that flow. The flow property implies that
$x_r=x_t+(r-t)u^v(s,x_t,r,t)$ is constant when differentiating along the same
trajectory with respect to its starting time $t$. Hence
\begin{equation}
0=v(s,x_t,t)-u^v(s,x_t,r,t)
+(r-t)\mathcal D_t^v u^v.
\end{equation}
Rearranging gives the identity used in the main text. The derivation holds for
$r>t$; the diagonal value follows from continuity as $r$ approaches $t$.
Holding $r$ fixed is essential: differentiating a destination that moves with $t$
would introduce an additional term.

\subsection{Conditional target calculation}
Fix the student parameters and the environment state. Assume the required
conditional moments exist, and let $m(x,t)=\E[\widehat v_t\mid X_t=x]$.
If $r$ is conditionally independent of the adjoint-label noise given $(X_t,t)$,
the derivatives of the student are deterministic under this conditioning.
Writing the numerical value of the detached target as $T$, linearity gives
\begin{align}
\E[T\mid x,r,t]
&=m(x,t)+(r-t)\big[\partial_tu_\theta\nonumber\\
&\hspace{32mm}+(\partial_xu_\theta)m(x,t)\big].
\end{align}
Setting $m=v^*$ proves Eq.~\eqref{eq:conditional}. Stop-gradient controls the
optimization derivative; it does not alter this equality of numerical values.
For a frozen square-integrable target, unweighted squared-error regression has
its conditional mean as a minimizer. Applying this observation to a target that
is recomputed from the student provides a compatibility statement, not a proof
that repeated training updates converge.

Residual-dependent weighting changes the regression stationarity condition to a
weighted residual equation. The preceding conditional-mean argument therefore
does not establish the optimizer of the practical $p=0.3$ objective. Nor does
this calculation prove the premise $m=v^*$ for our endpoint-noised trajectory law.
A transfer of an AM stationary-point result would additionally require its
path-measure and reference-field assumptions; the calculation here makes no
such transfer.

\subsection{Endpoint-conditioned path measures}
Suppose $P^\beta$ is an exact reference path measure with terminal law $p_\beta$,
and $q$ is absolutely continuous with respect to $p_\beta$. Define a path measure
$\widetilde P_q$ by drawing $X_1\sim q$ and then the reference conditional path.
Disintegration gives
\begin{equation}
\frac{d\widetilde P_q}{dP^\beta}(X)
=\frac{dq}{dp_\beta}(X_1).
\end{equation}
If the logarithm is integrable, taking expectation establishes
\begin{equation}
D_{\mathrm{KL}}(\widetilde P_q\Vert P^\beta)
=D_{\mathrm{KL}}(q\Vert p_\beta).
\end{equation}
If $q$ is the ideal terminal tilt proportional to $p_\beta e^{Q/\lambda}$,
this construction recovers the corresponding terminally tilted reference path
measure. Independence of reference endpoints preserves the prescribed initial
law. This does not identify the learned diagonal field as the drift of
$\widetilde P_q$, and therefore does not turn our measured velocity energy into
its exact KL. It also assumes an exact conditional kernel, not merely a learned
reference with approximately correct marginals.
\section{Implementation and Gradient Boundaries}
\subsection{Behavior MeanFlow training}
\label{app:behavior}
For a dataset action $a$ and independent $z\sim\mathcal N(0,I)$, sample
$t\sim\mathcal U(0,1)$ and set $x_t=(1-t)z+ta$, $c_\beta=a-z$.
We train the behavior MeanFlow at the three destinations in
Eq.~\eqref{eq:three-r} with the target
\begin{equation}
T_\beta=\sg\!\left[c_\beta+(r-t)\mathcal D_t^{c_\beta}
u_\beta(s,x_t,r,t)\right].\label{eq:behavior}
\end{equation}
Its regression uses the residual weighting of Eq.~\eqref{eq:weighted-loss}.
After 300k behavior updates, the live policy and its EMA are initialized from
the behavior network with a fresh optimizer state. During offline learning,
the behavior field continues to update on dataset actions. The behavior EMA
is fixed within each adjoint target construction but evolves across updates.
Only its diagonal velocity enters the adjoint reference; the behavior action
sampler also queries non-diagonal time pairs during two-call refinement.

For each offline update, the actor computation proceeds as follows:
\begin{enumerate}
\item Generate a two-call policy endpoint and the reverse Markov trajectory.
Detach the path from policy-parameter differentiation.
\item Differentiate the clipped-action target-critic mean with respect to its
action input, then propagate the negative terminal derivative backwards with
reference-drift VJPs as in Eq.~\eqref{eq:discrete-adjoint}.
\item Form $\widehat v_k$ using the regularized diffusion schedule and effective
$\lambda$. Freeze this label and sample three destination times per point.
\item Evaluate the live student's prediction and its input/time JVP at fixed
destination. Detach the complete target and residual-based weight.
\item Differentiate only the prediction branch of the regression objective.
Update the behavior model and critic using their own losses, update target
networks, and then update the controller using the measured surrogate.
\end{enumerate}
Freezing reference parameters must not disable their input Jacobians.
Likewise, retaining student input derivatives for target construction does not
mean differentiating through that target during optimization. The policy is
initialized from the live pretrained behavior network, with corresponding EMA
copies reset and a fresh optimizer state. No historical policy checkpoint is
used to substitute for the designated pretraining in a center run.

\begin{figure*}[!t]
\centering
\includegraphics[width=\textwidth]{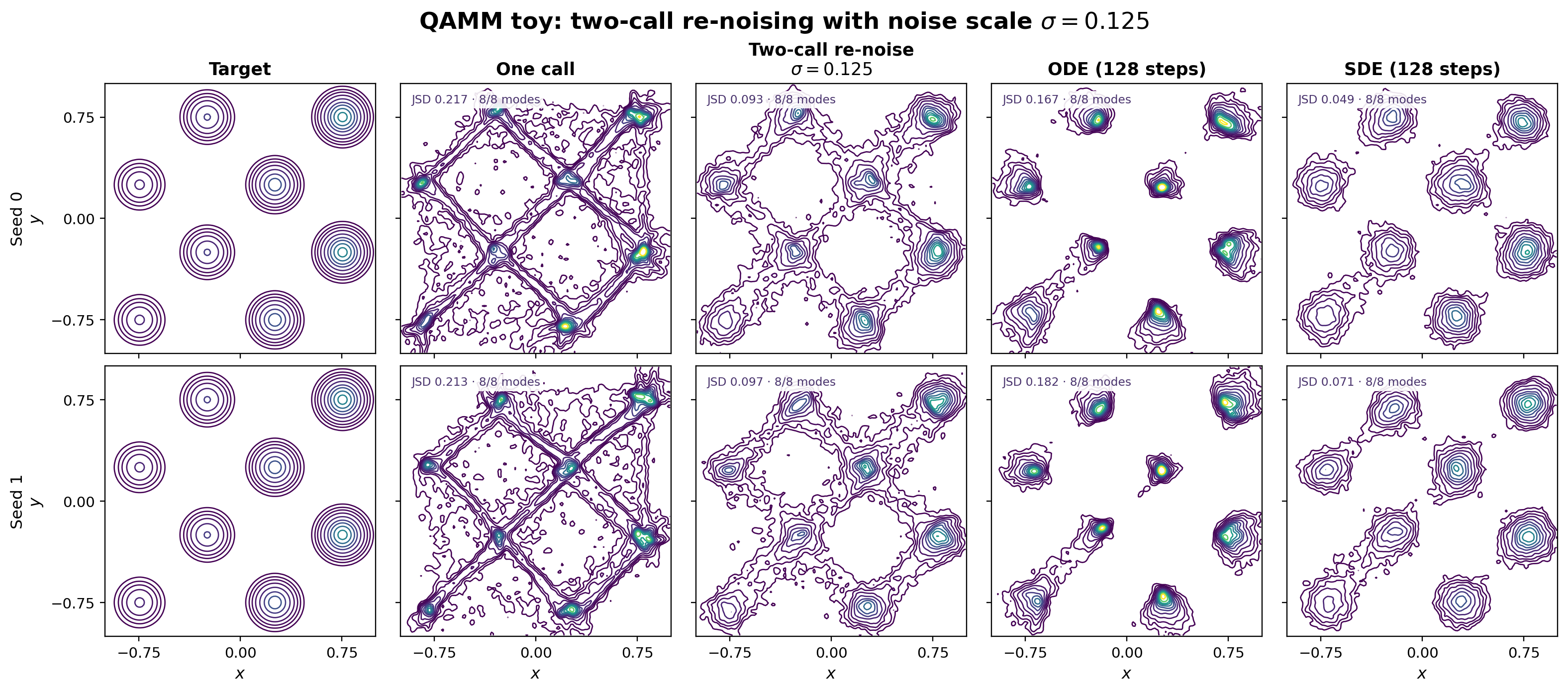}
\caption{Eight-mode QAMM toy diagnostic for two seeds (rows). Columns
show the target and generated densities with one call, two-call endpoint
re-noising at noise scale $\sigma=0.125$, 128-step ODE, and 128-step SDE
sampling. For the two-call sampler, $x_t=(1-\sigma)\widehat x_1+\sigma z$
with $z\sim\mathcal N(0,I)$, so the internal forward time is $t=0.875$.
Panel annotations give grid Jensen--Shannon divergence and covered modes.
This training protocol differs from the HumanoidMaze main experiments.}
\label{fig:qamm-toy}
\end{figure*}

\section{Configurations and Reproducibility}
\label{app:config}
\begin{table}[h]
\centering\small
\caption{Recorded Humanoid center configuration.}
\label{tab:config}
\begin{tabular}{lr}
\toprule
Setting & Value\\\midrule
Behavior pretraining / offline updates & 300k / 1M\\
Batch size / critic ensemble size & 256 / 10\\
Action horizon / discount & 1 / 0.999\\
Policy hidden layers & $512\times4$\\
Actor learning rate & $10^{-4}$\\
Behavior and critic learning rate & $3\times10^{-4}$\\
Training path points / deployment calls & 10 / 2\\
Refinement time & 0.5\\
Residual weighting $(p,c)$ & $(0.3,10^{-3})$\\
Backup mixture $\alpha$ / KL budget & 0.25 / 0.5\\
Effective initial $\lambda$ / scale $\kappa$ & 3 / 3\\
Controller step size / EMA coefficient & 0.01 / 0.1\\
Controller internal bounds & [0.01,100]\\
Evaluation interval / episodes & 50k / 50\\
\bottomrule
\end{tabular}
\end{table}
The center configuration disables additional direct-Q refinement, behavior-map
supervision, and energy-weighted auxiliary losses. The two training seeds are
10001 and 20002. Exploration of different $\alpha$ and KL budgets is reported
separately. Source snapshots and launcher manifests, rather than mutable agent
defaults alone, define a run.
Each run has 1M offline and 500k online updates. At every 50k updates we
evaluate 50 episodes. For each phase, table entries average its final five
evaluations within each seed and then report the mean and standard deviation
across seeds. The supplied TRQAM-author baseline logs contain eight seeds;
their implementations and evaluation randomness were not rerun under a
common harness. Large tuning changes $\alpha$ to $0.5$ and the KL budget to
$0.75$, selected on seed 10001 before running seed 20002.
The generic agent configuration contains historical defaults that differ from
the recorded settings in Table~\ref{tab:config}. Per-seed evaluation CSV files
and the plotting script accompany this draft in the \texttt{arxiv/data} and
\texttt{arxiv/plot\_humanoid\_curves.py} paths.

\section{Exploratory QAMM Toy Diagnostic}
\label{app:qamm-toy}
Figure~\ref{fig:qamm-toy} compares a prescribed eight-mode
$\exp(Q)$-tilted target with QAMM samples in a two-dimensional diagnostic.
Two independently initialized runs each use 5k flow-matching pretraining
updates, 5k QAMM updates, and 30k samples per sampler. We retain a single
two-call re-noising scale, $\sigma=0.125$ (internal $t=0.875$), selected
by a checkpoint-only time sweep. All samplers cover the eight modes. Grid
Jensen--Shannon divergence for the two-call sampler is $0.093$ and $0.097$
across the two seeds, compared with $0.049$ and $0.071$ for the 128-step
SDE reference. Some connecting mass remains and mode-weight total variation
is slightly worse than at $\sigma=0.5$; the figure therefore does not show
that few-call sampling reproduces the ideal tilted distribution.

This diagnostic uses a frozen flow-matching reference, a randomly
initialized actor, and one sampled destination per training example.
It differs from the HumanoidMaze procedure, which initializes from a
behavior MeanFlow, uses three destinations per path point, and constructs
an endpoint-noised Markov path. Accordingly, the figure probes a
simplified protocol rather than validating the full QAMM implementation.